\documentclass{ecai}
\usepackage{times}
\usepackage{graphicx}
\usepackage{latexsym}

\usepackage{amsmath}
\usepackage{graphicx}
\usepackage{algorithm}
\usepackage{hyperref}
\usepackage{algorithmic}
\usepackage{amsmath}
\usepackage{amssymb}
\usepackage{multirow}
\usepackage{booktabs}
\usepackage[numbers]{natbib}

\newcommand{\R}{\mathbb{R}}

\DeclareMathOperator{\sut}{subject\ to\ \ \ \ }
\DeclareMathOperator{\eps}{eps}

\usepackage{amsmath}
\usepackage{tikz}
\usepackage{mathdots}
\usepackage{yhmath}
\usepackage{cancel}
\usepackage{color}
\usepackage{siunitx}
\usepackage{array}
\usepackage{multirow}
\usepackage{amssymb}
\usepackage{gensymb}
\usepackage{tabularx}
\usepackage{booktabs}
\usetikzlibrary{fadings}

\usepackage{authblk}

\makeatletter
\def\blfootnote{\xdef\@thefnmark{}\@footnotetext}
\makeatother

\begin{document}

\title{A Lagrangian Approach to Information Propagation in Graph Neural Networks}

\author[1]{Matteo Tiezzi}
\author[1,2]{Giuseppe Marra}
\author[1]{Stefano Melacci}
\author[1]{ Marco Maggini}
\author[1]{ Marco Gori}
\affil[1]{Department of Information Engineering and Science (DIISM), University of Siena,
Siena, Italy\\
\{mtiezzi,mela,maggini,marco\}@diism.unisi.it 
\\}
\affil[2]{Department of Information Engineering (DINFO), University of Florence, 
Florence, Italy \\

g.marra@unifi.it}
\date{}                     
\setcounter{Maxaffil}{0}
\renewcommand\Affilfont{\itshape\small}

\maketitle
\bibliographystyle{ecai}

\maketitle

\begin{abstract}
In many real world applications, data are characterized by a complex structure, that can be naturally encoded as a graph. In the last years, the popularity of deep learning techniques has renewed the interest in neural models able to process complex patterns. In particular, inspired by the Graph Neural Network (GNN) model, different architectures have been proposed to extend the original GNN scheme. GNNs exploit a set of state variables, each assigned to a graph node, and a diffusion mechanism of the states among neighbor nodes, to implement an iterative procedure to compute the fixed point of the (learnable) state transition function.
In this paper, we propose a novel approach to the state computation and the learning algorithm for GNNs, based on a constraint optimisation task solved in the Lagrangian framework. The state convergence procedure is implicitly expressed by the constraint satisfaction mechanism and does not require a separate iterative phase for each epoch of the learning procedure. 
In fact, the computational structure is based on the search for saddle points of the Lagrangian in the adjoint space composed of weights, neural outputs (node states), and Lagrange multipliers. 
The proposed approach is compared experimentally with other popular models for processing graphs. 
\end{abstract}

\section{Introduction}
\label{sec:intro}
\blfootnote{Accepted for publication at the European Conference on Artificial Intelligence (ECAI) 2020 (DOI: TBA).
}

Due to their flexibility and approximation capabilities, the original processing and learning schemata of Neural Networks have been extended in order to deal with structured inputs. Based on the original feedforward model, able to process vectors of features as inputs, different architectures have been proposed to process sequences (Recurrent Neural Networks \cite{Williams1989ALA}), rasters of pixels (Convolutional Neural Networks \citep{LeCun:1998:CNI:303568.303704}), directed acyclic graphs (Recursive Neural Networks \citep{Goller96,Frasconi:1998:GFA:2325763.2326281}), and general graph structures (Graph Neural Networks \citep{DBLP:journals/tnn/ScarselliGTHM09}). All these models generally share the same learning mechanism based on the error BackPropagation (BP) through the network architecture, that allows the computation of the loss gradient with respect to the connection weights. When processing structured data the original BP schema is straightforwardly extended by the process of {\em unfolding} that generates a network topology based on the current input structure by replicating a base neural network module (e.g. BP Through Time, BP Through Structure).

However, recently, some works \citep{carreira2014distributed} proposed a different approach to learning neural networks, where neural computations are expressed as constraints and the optimization is framed into the Lagrangian framework. These algorithms are naturally local and allow the learning of any computational structure, both acyclical or cyclical. The main drawback of these methods is that they are quite memory inefficient; in particular, they need to keep an extra-variable for each hidden neuron and for each example. This makes them inapplicable to large problems where BP is still the only viable option. 

Graph Neural Networks (GNNs) \citep{DBLP:journals/tnn/ScarselliGTHM09} exploit neural networks to learn how to encode nodes of a graph for a given task taking into account both the information local to each node and the whole graph topology. The learning process requires, for each epoch, an iterative diffusion mechanism up to convergence to a stable fixed point, that is computationally heavy. A maximum number of iterations can be defined but this limits the local encoding to a maximum depth of the neighborhood of each node. In this paper, we propose a new learning mechanism for GNNs based on a Lagrangian formulation that allows the embedding of the fixed point computation into the problem constraints. In the proposed scheme the network state representations and the weights are jointly optimized without the need of applying the fixed point relaxation procedure at each weight update epoch. 

The paper is organized as follows. The next section reviews the main developments in both the Neural Network models for processing graphs and learning methods based on the Lagrangian approach. Section \ref{sec:gnn} introduces the basics of the GNN model, whereas in Section \ref{sec:Lgnn} the Lagrangian formulation of GNNs is described. Section \ref{sec:experiments} reports the experimental evaluation of the proposed constraint--based learning for GNNs. Finally, the conclusions are drawn in Section \ref{sec:concl}.

\section{Related Works}
\label{sec:related}

In many applications data are characterized by an underlying structure that lays on a non-Euclidean domain, i.e. graphs and manifolds. Whilst commonly addressed in relational learning, such domains have been initially not taken into account by popular machine learning techniques, that have been mostly devised for grid--like and Euclidean structured data \citep{DBLP:journals/spm/BronsteinBLSV17}. 
Early machine learning approaches for structured data were designed for directed acyclic graphs \citep{Sperduti:1997:SNN:2325755.2326105, Frasconi:1998:GFA:2325763.2326281}, while a more general framework was introduced in \citep{DBLP:journals/tnn/ScarselliGTHM09}. GNNs are able to directly deal with directed, undirected and cyclic graphs. The core idea is based on an iterative scheme of information diffusion among neighboring nodes, involving a propagation process aimed at reaching an equilibrium of the node states that represent a local encoding of the graph for a given task. The encoding is a computationally expensive process being based on the computation of the fixed point of the state transition function. Some proposals were aimed at simplifying this step, such as the scheme proposed in \citep{DBLP:journals/corr/LiTBZ15} that exploits gated recurrent units. 

Recent approaches differ in the choice of neighborhood aggregation method and graph level pooling scheme, and can be categorized in two main areas. \textit{Spectral approaches} exploit particular embeddings of the graph and the convolution operation defined in the spectral domain \citep{DBLP:journals/corr/BrunaZSL13}. However, they are characterized by computational drawbacks caused by the eigen--decomposition of the graph Laplacian. Simplified approaches are based on smooth reparametrization \citep{DBLP:journals/corr/HenaffBL15} or approximation of the spectral filters by a Chebyshev expansion \citep{DBLP:conf/nips/DefferrardBV16}. Finally, in Graph Convolutional Networks (GCNs) \citep{DBLP:conf/iclr/KipfW17}, filters are restricted to operate in a 1-hop neighborhood of each node. \textit{Spatial methods}, instead, exploit directly the graph topology, without the need of an intermediate representation. These approaches differ mainly in the definition of the aggregation operator used to compute the node states, that must be able to maintain weight sharing properties and to process nodes with different numbers of neighbors.
The PATCHY-SAN \citep{DBLP:conf/icml/NiepertAK16} model converts graph-structured data into a grid-structured representation,  extracting and normalizing neighborhoods containing a fixed number of nodes. In \citep{DBLP:conf/nips/DuvenaudMABHAA15} the model exploits a weight matrix for each node degree, whereas DCNNs \citep{DBLP:conf/nips/AtwoodT16} compute the hidden node representation convolving inputs channels with power series of the transition probability matrix, learning weights for each neighborhood degree. GraphSAGE \citep{hamilton2017inductive} exploits different aggregation functions to merge the node neighborhood information. Deep GNNs \citep{Bianchini2018} stack layers of GNNs to obtain a deep architecture. In the context of graph classification tasks, SortPooling \citep{DBLP:conf/aaai/ZhangCNC18} uses a framework based on DGCNNs with a pooling strategy, that performs pooling by ordering vertices. Finally,  the representational and discriminative power of GNN models were explored in \citep{DBLP:journals/corr/abs-1810-00826}, also introducing the novel GIN model.

A Lagrangian formulation of learning can be found in the seminal work of Yann LeCun \cite{lecun1988theoretical}, which studies  a theoretical framework for Backpropagation. More recently, Carreira and Wang \cite{carreira2014distributed} introduced the idea of training networks, transformed into a constraints-based representation, though an extension of the learnable parameters space. Their optimization scheme was based on quadratic penalties, aiming at an approximate solution of the problem afterwards refined by a post-processing phase.
Differently, \cite{taylor2016training} exploits closed-form solutions were most of the architectural constraints are softly enforced, and further additional variables are introduced to parametrize the neuron activations.

By framing the optimization of neural networks in the Lagrangian framework, where neural computations are expressed as constraints, their main goal is to obtain a local algorithm where computations of different layers can be carried out in parallel. On the contrary in the proposed approach, we use a novel mixed strategy. In particular, the majority of the computations still rely on Backpropagation while constraints are exploited only to express the diffusion mechanism. This allows to carry out both the optimization of the neural functions and the diffusion process at the same time, instead of alternating them into two distinct phases (as in \cite{DBLP:journals/tnn/ScarselliGTHM09}), with a theoretical framework supporting this approach (Lagrangian optimization).

It has already been shown that algorithms on graphs can be effectively learned exploiting a constrained fixed-point formulation. For example, SSE \cite{dai2018learning} exploits the Reinforcement Learning \textit{policy iteration} algorithm for the interleaved evaluation of the fixed point equation and the improvement of the transition and output functions.
Our approach, starting from similar assumptions, exploits the unifying Lagrangian framework for learning both the transition and the output functions. Thus, by framing the optimization algorithm into a standard gradient descent/ascent scheme, we are allowed to use recent update rules (e.g. Adam) without the need to resort to ad-hoc moving average updates.\\

\section{Graph Neural Networks}
\label{sec:gnn}

The term Graph Neural Network (GNN) refers to a general computational model, that exploits the processing and learning schemes of neural networks to process non Euclidean data, i.e. data organized as graphs.

Given an input graph $G=(V,E)$, where $V$ is a finite set of {\em nodes} and $E \subseteq V \times V$ collects the {\em arcs}, GNNs apply a two-phase computation on $G$. In the \textit{encoding} (or {\em aggregation}) phase the model computes a state vector for each node in $V$ by (iteratively) combining the states of neighboring nodes (i.e. nodes $u, v \in V$ that are connected by an arc $(u,v) \in E$). In the second phase, usually referred to as \textit{output} (or {\em readout}), the latent representations encoded by the states stored in each node are exploited to compute the model output. The GNN can implement either a \textit{node-focused} function, where an output is produced for each node of the input graph, or a \textit{graph-focused} function, where the representations of all the nodes are aggregated to yield a single output for the whole input graph.

The GNN is defined by a pair of (learnable) functions, that respectively implement the {\em state transition} function $f_a$ required in the encoding phase and the {\em output} function $f_r$ exploited in the output phase, as follows:
\begin{align}
        x_v^{(t)} &= f_a(x_{{\mbox{\scriptsize ne}}[v]}^{(t-1)}, l_{{\mbox{\scriptsize ne}}[v]}, l_{(v,{\mbox{\scriptsize ch}}[v])},l_{({\mbox{\scriptsize pa}}[v],v)}, x_{v}^{(t-1)}, l_{v}|\theta_{f_a}),
        \label{eq:aggregation}
\end{align}
\begin{subequations}
\begin{align}
  \label{eq:readout_node}  y_v \ &= f_r(x_v^{(T)}|\theta_{f_r}),\\
  \label{eq:readout_graph} y_G \ &= f_r(\{x_v^{(T)}, v \in V\}|\theta_{f_r}),
\end{align}
\end{subequations}
where $x_v^{(t)} \in \R^s$ is the state of the node $v$ at iteration $t$, $\mbox{pa}[v] = \{u \in V: (u,v) \in E \}$ is the set of the {\em parents} of node $v$ in $G$,  $\mbox{ch}[v] = \{u \in V: (v,u) \in E \}$ are the {\em children} of $v$ in $G$, $\mbox{ne}[v]=\mbox{pa}[v] \cup\mbox{ch}[v]$ are the {\em neighbors} of the node $v$ in $G$, $l_u \in \R^m$ is the feature vector available for node $u \in V$, and $l_{(u,w)} \in \R^d$ is the feature vector available for the arc $(u,w) \in E$\footnote{With abuse of notation, we denote the set $\{x_u^{(t-1)}: u \in {\mbox{ne}}[v]\}$ by $x_{{\mbox{\scriptsize ne}}[v]}^{(t-1)}$. Similar definitions apply for $l_{{\mbox{\scriptsize ne}}[v]}$, $l_{(v,{\mbox{\scriptsize ch}}[v])}$, and $l_{({\mbox{\scriptsize pa}}[v],v)}$.}. The vectors $\theta_{f_a}$ and $\theta_{f_r}$ collect the model parameters (the neural network weights) to be adapted during the learning procedure. Equations (\ref{eq:readout_node}) and (\ref{eq:readout_graph}) are the two variants of the output function for node-focused or graph-focused tasks, respectively.

\begin{table*}
    \centering
     
    \begin{tabular}{llc}
        \toprule
        Method: Function & Reference & Implementation of $f_a$\\
         \midrule
         GNN: Sum & Scarselli et al. \cite{DBLP:journals/tnn/ScarselliGTHM09}   &  $\sum_{u \in \mbox{\scriptsize ne}[v]} h(x_u,l_u,l_{(v,u)},l_{(u,v)},x_v,l_v|\theta_h)$\\
         GIN: Sum  & Xu et al. \cite{DBLP:journals/corr/abs-1810-00826} & $h(x_v + \sum_{u \in \mbox{\scriptsize ne}[v]} x_u$) \\
         GCN: Mean & Kipf and Welling \cite{DBLP:conf/iclr/KipfW17} & $h\bigg (\frac{1}{|\mbox{\scriptsize ne}[v]| + 1} (x_v + \sum_{u \in \mbox{\scriptsize ne}[v]} x_u$)\bigg)\\
         GraphSAGE: Max & Hamilton et al. \cite{hamilton2017inductive} & $\max_{u \in \mbox{\scriptsize ne}[v]} h(x_u)$ \\
         \bottomrule\\
    \end{tabular}
   \caption{Simplified implementations of the state transition function $f_a$. The function $h()$ is implemented by a feedforward neural network with $s$ outputs, whose input is the concatenation of its arguments (f.i. in the first case the input consists of a vector of $2s+2m+2d$ entries, with $l_{(u,v)} \in \R^d$ and  $l_u \in \R^m$). For the sake of clarity, some of these formulas are reported in a simplified way w.r.t. the original proposal. For example,  the "mean" function in \cite{DBLP:conf/iclr/KipfW17} is a weighted mean, where the weights come from the normalized graph adjacency matrix, or the "max" function in   \cite{hamilton2017inductive} is followed by a concatenation.}
    \label{tab:gnn}
\end{table*}
In Table~\ref{tab:gnn}, we show some possible choices of the function $f_a$. It should be noted that this function may depend on a variable number of inputs, given that the nodes $v \in V$ may have different degrees $\mbox{de}[v]=|\mbox{ne}[v]|$. Moreover, in general, the proposed implementations are invariant with respect to permutations of the nodes in $\mbox{ne}[v]$, unless some predefined ordering is given for the neighbors of each node.

$T$ is the number of iterations of the state transition function applied before computing the output. The recursive application of the state transition function $f_a$ on the graph nodes yields a diffusion mechanism, whose range depends on $T$. In fact, by stacking $t$ times the aggregation of 1-hop neighborhoods by $f_a$, information of one node can be transferred to the nodes that are distant at most $t$-hops. The number $t$ may be seen as the \textit{depth} of the GNN and thus each iteration can be considered a different \textit{layer} of the GNN. A sufficient number of layers is key to achieve a useful encoding of the input graph for the task at hand and, hence, the choice is problem--specific.

In the original GNN model \citep{DBLP:journals/tnn/ScarselliGTHM09}, eq. (\ref{eq:aggregation}) is executed until convergence of the state representation, i.e. until $x_v^{(t)} \simeq x_v^{(t-1)}, v \in V$. This scheme corresponds to the computation of the {\em fixed point} of the state transition function $f_a$ on the input graph. In order to guarantee the convergence of this phase, the transition function is required to be a {\em contraction map}.\\
Henceforth,  for compactness, we denote the state transition function, applied to a node  $v \in V$,  with:
\begin{equation}
f_{a,v} = f_a(x_{{\mbox{\scriptsize ne}}[v]}, l_{{\mbox{\scriptsize ne}}[v]}, l_{(v,{\mbox{\scriptsize ch}}[v])},l_{({\mbox{\scriptsize pa}}[v],v)}, x_{v}, l_{v}|\theta_{f_a}).
\end{equation}

Basically, the encoding phase, through the iteration of $f_a$, finds a solution to the fixed point problem defined by the constraint
\begin{equation}
    \forall v \in V, x_v = f_{a,v}
    \label{eq:fixedpoint}
\end{equation}
In this case, the states encode the information contained in the whole graph. This diffusion mechanism is more general than executing only a fixed number of iterations (i.e. stacking a fixed number of layers). However, it can be computationally heavy and, hence, many recent GNN architectures apply only a fixed number of iterations for all nodes.

\section{A constraint-based formulation of Graph Neural Networks}
\label{sec:Lgnn}
Neural network learning can be cast as a Lagrangian optimization problem by a formulation that requires the minimization of the classical data fitting loss (and eventually a regularization term) and the satisfaction of a set of {\em architectural} constraints that describe the computation performed on the data. Given this formulation, the solution can be computed by finding the {\em saddle points} of the associated Lagrangian in the space defined by the original network parameters and the {\em Lagrange multipliers}. The constraints can be exploited to enforce the computational structure that characterizes the GNN models.

The computation of Graph Neural Networks is driven by the input graph topology that defines the constraints among the computed state variables $x_v, v \in V$. In particular, the fixed point computation aims to solving eq.~(\ref{eq:fixedpoint}), that imposes a constraint between the node states and the way they are computed by the state transition function. \\
In the original GNN learning algorithm, the computation of the fixed point is required at each epoch of the learning procedure, as implemented by the iterative application of the transition function. Moreover, also the gradient computation requires us to take into account the relaxation procedure, by a backpropagation schema through the replicas of the state transition network exploited during the iterations for the fixed point computation. This procedure may be time consuming when the number of iterations $T$ for convergence to the fixed point is high (for instance in the case of large graphs).

We consider a Lagrangian formulation of the problem by adding free variables corresponding to the node states $x_v$, such that the fixed point is directly defined by the constraints themselves, as
\begin{align}
\forall v \in V, \ {\mathcal G}\left(x_v - f_{a,v}\right)=0
\label{eq:constraint}
\end{align}
where ${\mathcal G}(x)$ is a function characterized by $\mathcal{G}(0)=0$, such that the satisfaction of the constraints implies the solution of eq.~(\ref{eq:fixedpoint}).
Apart from classical choices, like $\mathcal{G}(x) = x$ or   $\mathcal{G}(x) = x^2$, we can design different function shape (see Section~\ref{sec:artificial}), with desired properties. For instance, a possible implementations is ${\mathcal G}(x)=\max(||x||_1-\epsilon,0)$, where $\epsilon \geq 0$ is a parameter that can be used to allow tolerance in the satisfaction of the constraint. The hard formulation of the problem requires $\epsilon=0$, but by setting $\epsilon$ to a small positive value it is possible to obtain a better generalization and tolerance to noise.

In the following, for simplicity, we will refer to a node-focused task, such that for some (or all) nodes $v \in S \subseteq V$ of the input graph $G$, a target output $y_v$ is provided as a supervision\footnote{For the sake of simplicity we consider only the case when a single graph is provided for learning. The extension for more graphs is straightforward for node-focused tasks, since they can be considered as a single graph composed by the given graphs as disconnected components.}. If $L(f_r(x_v|\theta_{fr_r}),y_v)$ is the loss function used to measure the target fitting approximation for node $v \in S$, the formulation of the learning task is: 
\begin{align}
 \nonumber \min_{\theta_{f_a},\theta_{f_r}, X} \ \ \ \ & \sum_{v \in S} L(f_r(x_v|\theta_{f_r}), y_v)\\
\sut & \mathcal{G}\left(x_v - f_{a,v}\right) = 0, \quad \forall \ v \in V
\label{eq:main_problem_1}
\end{align}
where $\theta_{f_a}$ and $\theta_{f_r}$ are the weights of the MLPs implementing the state transition function and the output function, respectively, and $X = \{x_v: v \in V\}$ is the set of the introduced free state variables.

This problem statement implicitly includes the definition of the fixed point of the state transition function in the optimal solution, since for any solution the constraints are satisfied and hence the computed optimal $x_v$ are solutions of eq.~(\ref{eq:fixedpoint}).
As shown in the previous subsection, the constrained optimization problem of eq.~(\ref{eq:main_problem_1}) can be faced in the Lagrangian framework by introducing for each constraint a Lagrange multiplier $\lambda_v$, to define the Lagrangian function $\mathcal{L}(\theta_{f_a},\theta_{f_r}, X, \Lambda)$ as:
\begin{multline}
\mathcal{L}(\theta_{f_a},\theta_{f_r}, X, \Lambda) = 
\sum_{v \in S} \left[ L(f_r(x_v|\theta_{f_r}), y_v) + \right. \\
\left. + \lambda_v \mathcal{G}\left(x_v - f_{a,v}\right)\right],
\label{eq:GNNlagrangian}
\end{multline}
where $\Lambda$ is the set of the $|V|$ Lagrangian multipliers. Finally, we can define the unconstrained optimization problem as the search for saddle points in the adjoint space 
$(\theta_{f_a},\theta_{f_r}, X, \Lambda)$ as:
\begin{align}
 \underset{\theta_{f_a},\theta_{f_r}, X}{\min}   \ \underset{\Lambda}{\max}\ \  \mathcal{L}(\theta_{f_a},\theta_{f_r}, X, \Lambda)
\label{eq:final}
\end{align}
that can be solved by gradient descent with respect to the variables $\theta_{f_a},\theta_{f_r}, X$ and gradient ascent with respect to the Lagrange multipliers $\Lambda$, exploiting the Basic Differential Multiplier Method, introduced in \cite{Platt} in the context of neural networks. We are interested in having a strong enforcement of the \textit{diffusion constraints}, and common penalty-based methods are hard to tune and not always guaranteed to converge to the constraint satisfaction. BDMM could be seen as a simplified procedure that implements the principles behind the common Multiplier Methods, in order to enforce the hard fulfilment of the given constraints.\\
The gradient can be computed locally to each node, given the local variables and those of the neighboring nodes. In fact, the derivatives of the Lagrangian \footnote{When parameters are vectors, the reported gradients should be considered element-wise.} with respect to the considered parameters are:

\begin{align}
\frac{\partial \mathcal{L}}{\partial x_v} & = L'f'_{r,v} + \lambda_v \mathcal{G}'_v (1 - f'_{a,v}) - \hskip-2mm \sum_{w:v \in ne[w]} \hskip-2mm \lambda_w \mathcal{G}'_w f'_{a,w} \\
\frac{\partial \mathcal{L}}{\partial \theta_{f_a}} & = - \sum_{v \in S} \lambda_v \mathcal{G}'_v f'_{a,v} \\
\frac{\partial \mathcal{L}}{\partial \theta_{f_r}} & = \sum_{v \in S} L'f'_{r,v} \\
\frac{\partial \mathcal{L}}{\partial \lambda_v} & = \mathcal{G}_v 
\end{align}
where, $f_{a,v} = f_a(x_{{\mbox{\scriptsize ne}}[v]}, l_{{\mbox{\scriptsize ne}}[v]}, l_{(v,{\mbox{\scriptsize ch}}[v])},l_{({\mbox{\scriptsize pa}}[v],v)}, x_{v}, l_{v}|\theta_{f_a})$, $f'_{a,v}$ is its first derivative\footnote{The derivative is computed with respect to the same argument as in the partial derivative on the left side.}, $f_{r,v} = f_r(x_v|\theta_{f_r})$, $f'_{r,v}$ is its first derivative, $\mathcal{G}_v = \mathcal{G}\left(x_v -f_{a,v}\right)$ and $\mathcal{G}'_v$ is its first derivative, and, finally, $L'$ is the first derivative of $L$. Being $f_{a}$ and $f_{r}$ implemented by feedforward neural networks, their derivatives are obtained easily by applying a classical backpropagation scheme, in order to optimize the Lagrangian function in the descent-ascent scheme, aiming at the saddle point, following \cite{Platt}.\\
We initialize the variables in $\mathcal{X}$ and $\Lambda$ to zero, while the neural weights $\theta_{f_a},\theta_{f_r}$ are randomly chosen.
In particular, this \textit{differential optimization} process consists of a gradient-descent step to update $\theta_{f_a},\theta_{f_r}, X$, and a gradient-ascent step to update $\Lambda$, until we converge to the desired stationary point. Hence, the redefined differential equation system gradually fulfills the constraints, undergoing oscillations along the constraint subspace. To ease this procedure, we add the $\mathcal{G}()$ function, with the purpose of obtaining a more stable learning process.

Even if the proposed formulation adds the free state variables $x_v$ and the Lagrange multipliers $\lambda_v$, $v \in V$, there is no significant increase in the memory requirements since the state variables are also required in the original formulation and there is just a Lagrange multiplier for each node.

The diffusion mechanism of the state computation is enforced by means of the constraints. The learning algorithm is based on a mixed strategy where \textit{(i)} Backpropagation is used to efficiently update the weights of the neural networks that implement the state transition and output functions, and, \textit{(ii)} the diffusion mechanism evolves gradually by enforcing the convergence of the state transition function to a fixed point by virtue of the constraints. This last point is a novel approach in training Graph Neural Networks. In fact, in classical approaches, the encoding phase (see Section~\ref{sec:gnn}) is completely executed during the forward pass to compute the node states and, only after this phase is completed, the backward step is applied to update the weights of $f_a$ and $f_r$. In the proposed scheme, both the neural network weights and the node state variables are simultaneously updated, forcing the state representation function towards a fixed point of $f_a$ in order to satisfy the constraints. In other words, the learning proceeds by jointly updating the function weights and by diffusing information among nodes, through their state, up to a stationary condition where both the objective function is minimized and the state transition function has reached a fixed point.\\
In our proposed algorithm, the diffusion process is turned itself into an optimization process that must be carried out both when learning and when making predictions.
As a matter of fact, inference itself requires the diffusion of information through the graph, that, in our case, corresponds with satisfying the constraints of Eq.~(\ref{eq:constraint}). For this reason, the testing phase requires a (short) optimization routine to be carried out, that simply looks for the satisfaction of Eq.~(\ref{eq:constraint}) for test nodes, and it is implemented using the same code that is used to optimize Eq.(\ref{eq:final}), avoiding to update the previously learned state transition and output functions. 

\subsection{Complexity analysis} 
Common graph models exploit synchronous updates among all nodes and multiple iterations for the node state embedding, with a computational complexity for each parameter update $\mathcal{O}(T(|V|+ |E|))$, where $T$ is the number of iterations, $|V|$ the number of nodes and $|E|$ the number of edges. By simultaneously carrying on the optimization of neural models and the diffusion process, our scheme relies only on 1-hop neighbors for each parameter update, hence showing a computational cost of $\mathcal{O}(|V|+ |E|)$. From the memory cost viewpoint, the persistent state variable matrix requires $\mathcal{O}(|V|)$ space.
However, it represents a much cheaper cost than most of GNN models, usually requiring $\mathcal{O}(T|V|)$ space. In fact, those methods need to store all the intermediate state values of all the iterations, for a latter use in back-propagation.
\section{Experiments}
\label{sec:experiments}

The evaluation was carried out on two classes of tasks. Artificial tasks (Subgraph matching and Clique detection) are commonly exploited as benchmarks for GNNs, thus, allowing a direct comparison of the proposed constraint based optimization algorithm with respect to the original GNN learning scheme, on the same architecture.
The second class of tasks consists of graph classification in the domains of social networks and bioinformatics. The goal is to compare the performances of the proposed approach, hereafter referred to as Lagrangian Propagation GNN (LP-GNN), that is based on a simpler model, with respect to deeper architectures such as Graph Convolutional Neural Networks.

With reference to Table \ref{tab:gnn}, in our experiments we validated two formulations of the state transition function $f_{a,v}$, with two different aggregation scheme.
In particular:
\begin{align}
 f_{a,v}^{\text{(SUM)}} & = \sum_{u \in ne[v]} h(x_u,l_u,l_{(v,u)},l_{(u,v)},x_v,l_v|\theta_h) \\
     f_{a,v}^{\text{(AVG)}} & = \tfrac{1}{|ne[v]|}\hskip-2mm \sum_{u \in ne[v]} \hskip-3mm h(x_u,l_u,l_{(v,u)},l_{(u,v)},x_v,l_v|\theta_h)
\end{align}

\subsection{Artificial Tasks}
\label{sec:artificial}

\begin{table*}[th!]
    \centering
    \begin{tabular}{c|c|c|c|c|c}
    \toprule

         & \textit{lin} & \textit{lin-$\epsilon$} & \textit{abs} & \textit{abs}-$\epsilon$ & \textit{squared}\\
         \toprule
           $\mathcal{G}(x)$ & $x$ & $\max(x, \epsilon) - \max(-x, \epsilon)$ &  $|x|$ & $\max(|x|- \epsilon, 0)$ & $x^2$ \\
          Unilateral & $\times$ &$\times$ & $\checkmark$ & $\checkmark$ & $\checkmark$\\
         $\epsilon$-insensitive & $\times$ & $\checkmark$ & $\times$ & $\checkmark$ & $\times$\\
         \bottomrule
    \end{tabular}
  
    \caption{The considered variants of the  $\mathcal{G}$ function. By introducing $\epsilon$-insensitive constraint satisfaction, we can inject into our hard-optimization scheme a controlled amount (i.e. $\epsilon$) of unsatisfaction tolerance. }
      \label{tab:g_functions}
\end{table*}

\paragraph{\textbf{Subgraph Matching}}

\begin{figure}[!ht]
\centering

\tikzset{every picture/.style={line width=0.75pt}} 

\begin{tikzpicture}[x=0.35pt,y=0.35pt,yscale=-1,xscale=1]

\draw  [color={rgb, 255:red, 0; green, 0; blue, 0 }  ,draw opacity=0 ][fill={rgb, 255:red, 23; green, 76; blue, 208 }  ,fill opacity=1 ] (135,197.64) .. controls (135,184.73) and (145.59,174.27) .. (158.65,174.27) .. controls (171.71,174.27) and (182.3,184.73) .. (182.3,197.64) .. controls (182.3,210.54) and (171.71,221) .. (158.65,221) .. controls (145.59,221) and (135,210.54) .. (135,197.64) -- cycle ;
\draw  [color={rgb, 255:red, 0; green, 0; blue, 0 }  ,draw opacity=0 ][fill={rgb, 255:red, 23; green, 76; blue, 208 }  ,fill opacity=1 ] (205,260.64) .. controls (205,247.73) and (215.59,237.27) .. (228.65,237.27) .. controls (241.71,237.27) and (252.3,247.73) .. (252.3,260.64) .. controls (252.3,273.54) and (241.71,284) .. (228.65,284) .. controls (215.59,284) and (205,273.54) .. (205,260.64) -- cycle ;
\draw  [color={rgb, 255:red, 0; green, 0; blue, 0 }  ,draw opacity=0 ][fill={rgb, 255:red, 23; green, 76; blue, 208 }  ,fill opacity=1 ] (65,258.64) .. controls (65,245.73) and (75.59,235.27) .. (88.65,235.27) .. controls (101.71,235.27) and (112.3,245.73) .. (112.3,258.64) .. controls (112.3,271.54) and (101.71,282) .. (88.65,282) .. controls (75.59,282) and (65,271.54) .. (65,258.64) -- cycle ;
\draw  [color={rgb, 255:red, 0; green, 0; blue, 0 }  ,draw opacity=0 ][fill={rgb, 255:red, 23; green, 76; blue, 208 }  ,fill opacity=1 ] (136,323.64) .. controls (136,310.73) and (146.59,300.27) .. (159.65,300.27) .. controls (172.71,300.27) and (183.3,310.73) .. (183.3,323.64) .. controls (183.3,336.54) and (172.71,347) .. (159.65,347) .. controls (146.59,347) and (136,336.54) .. (136,323.64) -- cycle ;
\draw    (174.9,212.84) -- (211.9,244.84) ;

\draw    (103.9,276.84) -- (140.9,308.84) ;

\draw    (178,308.28) -- (213.05,278.3) ;

\draw    (106,243.28) -- (142.05,213.3) ;

\draw  [color={rgb, 255:red, 0; green, 0; blue, 0 }  ,draw opacity=0 ][fill={rgb, 255:red, 23; green, 76; blue, 208 }  ,fill opacity=1 ] (318,44.64) .. controls (318,31.73) and (328.59,21.27) .. (341.65,21.27) .. controls (354.71,21.27) and (365.3,31.73) .. (365.3,44.64) .. controls (365.3,57.54) and (354.71,68) .. (341.65,68) .. controls (328.59,68) and (318,57.54) .. (318,44.64) -- cycle ;
\draw  [color={rgb, 255:red, 0; green, 0; blue, 0 }  ,draw opacity=0 ][fill={rgb, 255:red, 23; green, 76; blue, 208 }  ,fill opacity=1 ] (388,107.64) .. controls (388,94.73) and (398.59,84.27) .. (411.65,84.27) .. controls (424.71,84.27) and (435.3,94.73) .. (435.3,107.64) .. controls (435.3,120.54) and (424.71,131) .. (411.65,131) .. controls (398.59,131) and (388,120.54) .. (388,107.64) -- cycle ;
\draw  [color={rgb, 255:red, 0; green, 0; blue, 0 }  ,draw opacity=0 ][fill={rgb, 255:red, 23; green, 76; blue, 208 }  ,fill opacity=1 ] (248,105.64) .. controls (248,92.73) and (258.59,82.27) .. (271.65,82.27) .. controls (284.71,82.27) and (295.3,92.73) .. (295.3,105.64) .. controls (295.3,118.54) and (284.71,129) .. (271.65,129) .. controls (258.59,129) and (248,118.54) .. (248,105.64) -- cycle ;
\draw  [color={rgb, 255:red, 0; green, 0; blue, 0 }  ,draw opacity=0 ][fill={rgb, 255:red, 23; green, 76; blue, 208 }  ,fill opacity=1 ] (319,170.64) .. controls (319,157.73) and (329.59,147.27) .. (342.65,147.27) .. controls (355.71,147.27) and (366.3,157.73) .. (366.3,170.64) .. controls (366.3,183.54) and (355.71,194) .. (342.65,194) .. controls (329.59,194) and (319,183.54) .. (319,170.64) -- cycle ;
\draw    (357.9,59.84) -- (394.9,91.84) ;

\draw    (286.9,123.84) -- (323.9,155.84) ;

\draw    (361,155.28) -- (396.05,125.3) ;

\draw    (289,90.28) -- (325.05,60.3) ;

\draw    (435.3,107.64) -- (500.9,107.84) ;

\draw  [color={rgb, 255:red, 0; green, 0; blue, 0 }  ,draw opacity=0 ][fill={rgb, 255:red, 196; green, 11; blue, 18 }  ,fill opacity=1 ] (500.9,107.84) .. controls (500.9,94.94) and (511.49,84.48) .. (524.55,84.48) .. controls (537.61,84.48) and (548.2,94.94) .. (548.2,107.84) .. controls (548.2,120.74) and (537.61,131.2) .. (524.55,131.2) .. controls (511.49,131.2) and (500.9,120.74) .. (500.9,107.84) -- cycle ;
\draw  [color={rgb, 255:red, 0; green, 0; blue, 0 }  ,draw opacity=0 ][fill={rgb, 255:red, 196; green, 11; blue, 18 }  ,fill opacity=1 ] (402.3,236.64) .. controls (402.3,223.73) and (412.89,213.27) .. (425.95,213.27) .. controls (439.01,213.27) and (449.6,223.73) .. (449.6,236.64) .. controls (449.6,249.54) and (439.01,260) .. (425.95,260) .. controls (412.89,260) and (402.3,249.54) .. (402.3,236.64) -- cycle ;
\draw  [color={rgb, 255:red, 0; green, 0; blue, 0 }  ,draw opacity=0 ][fill={rgb, 255:red, 196; green, 11; blue, 18 }  ,fill opacity=1 ] (421.3,29.64) .. controls (421.3,16.73) and (431.89,6.27) .. (444.95,6.27) .. controls (458.01,6.27) and (468.6,16.73) .. (468.6,29.64) .. controls (468.6,42.54) and (458.01,53) .. (444.95,53) .. controls (431.89,53) and (421.3,42.54) .. (421.3,29.64) -- cycle ;
\draw  [color={rgb, 255:red, 0; green, 0; blue, 0 }  ,draw opacity=0 ][fill={rgb, 255:red, 196; green, 11; blue, 18 }  ,fill opacity=1 ] (157.3,61.64) .. controls (157.3,48.73) and (167.89,38.27) .. (180.95,38.27) .. controls (194.01,38.27) and (204.6,48.73) .. (204.6,61.64) .. controls (204.6,74.54) and (194.01,85) .. (180.95,85) .. controls (167.89,85) and (157.3,74.54) .. (157.3,61.64) -- cycle ;
\draw    (202.9,71.84) -- (250.9,96.84) ;

\draw    (362.9,35.84) -- (421.3,29.64) ;

\draw    (360.9,184.84) -- (408.9,219.84) ;

\draw (158,369) node  [align=left] {Target Subgraph};

\end{tikzpicture}

\caption{An example of a subgraph matching problem, where the graph with the blue nodes is matched against the bigger graph. }

\label{fig:subgraph}
\end{figure}
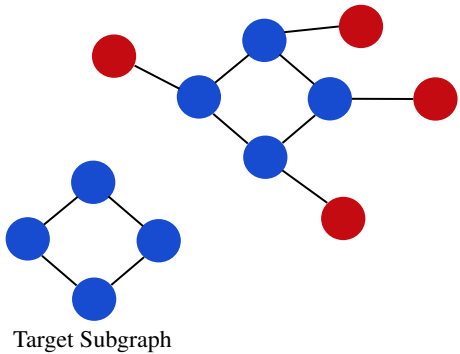

Given a graph $G$ and a graph $S$ such that $|S| \le |G|$, the subgraph matching problem consists in finding the nodes of a subgraph $\hat S \subset G$ which is isomorphic to $S$.  
The task is that of learning a function $\tau$, such that $\tau_S(G,n)=1, n \in V$, when the node $n$ belongs to the given subgraph $S$, otherwise $\tau_S(G,n)=0$.
It is designed to identify the nodes in the input graph that belong to a single subgraph given a priori during learning.
The problem of finding a given subgraph is common in many practical problems and corresponds, for instance, to finding a particular small molecule inside a greater compound.  An example of a subgraph structure is shown in Fig. \ref{fig:subgraph}.
Our dataset is composed of 100 different graphs, each one having 7 nodes. The number of nodes of the target subgraph $S$ is instead 3.
%

\begin{table*}[th!]

    \centering

\begin{tabular}{ccc|cc|cc}
\toprule
\multirow{2}{*}{Model} & & & \multicolumn{2}{c}{Subgraph} & \multicolumn{2}{c}{Clique} \\
\cmidrule{2-7}
& $\mathcal{G}$ & $\epsilon$ &    \textit{Acc(avg)}       &   \textit{Acc(std)}  & \textit{Acc(avg)}       &   \textit{Acc(std)} \\
\midrule

\multirow{7}{*}{LP-GNN} & \multirow{3}{*}{\textit{abs}} & 0.00 &  96.25 &  0.96  &  88.80 &  4.82 \\
 &  & 0.01 &  \textbf{96.30} &  0.87 &  88.75 &  5.03\\
 &  & 0.10 &  95.80 &  0.85 &  85.88 &  4.13 \\
  \cmidrule{2-7}
 & \multirow{3}{*}{\textit{lin}}  & 0.00 &  95.94 & 0.91 &   84.61 &  2.49 \\
 & & 0.01 &  95.94 &  0.91 &  85.21 &  0.54 \\
 & & 0.10 &  95.80 &  0.85 &  85.14 &  2.17 \\
  \cmidrule{2-7}
 & \textit{squared} & - &  96.17 &  1.01 &   \textbf{93.07} &  2.18 \\
 \midrule
GNN \cite{DBLP:journals/tnn/ScarselliGTHM09} & - & - & 95.86 &  0.64 &  91.86 &  1.12\\
\bottomrule

\end{tabular}

 \caption{Accuracies on the artificial datasets, for the proposed model (Lagrangian Propagation GNN - LP-GNN) and the standard GNN model for different settings.}
   \label{tab:artificial_results}
\end{table*}

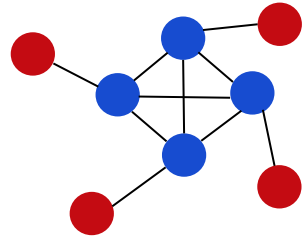
\begin{figure}[!ht]
\centering

\tikzset{every picture/.style={line width=0.75pt}} 

\begin{tikzpicture}[x=0.35pt,y=0.35pt,yscale=-1,xscale=1]

\draw  [color={rgb, 255:red, 0; green, 0; blue, 0 }  ,draw opacity=0 ][fill={rgb, 255:red, 23; green, 76; blue, 208 }  ,fill opacity=1 ] (318,44.64) .. controls (318,31.73) and (328.59,21.27) .. (341.65,21.27) .. controls (354.71,21.27) and (365.3,31.73) .. (365.3,44.64) .. controls (365.3,57.54) and (354.71,68) .. (341.65,68) .. controls (328.59,68) and (318,57.54) .. (318,44.64) -- cycle ;
\draw  [color={rgb, 255:red, 0; green, 0; blue, 0 }  ,draw opacity=0 ][fill={rgb, 255:red, 23; green, 76; blue, 208 }  ,fill opacity=1 ] (392,103.64) .. controls (392,90.73) and (402.59,80.27) .. (415.65,80.27) .. controls (428.71,80.27) and (439.3,90.73) .. (439.3,103.64) .. controls (439.3,116.54) and (428.71,127) .. (415.65,127) .. controls (402.59,127) and (392,116.54) .. (392,103.64) -- cycle ;
\draw  [color={rgb, 255:red, 0; green, 0; blue, 0 }  ,draw opacity=0 ][fill={rgb, 255:red, 23; green, 76; blue, 208 }  ,fill opacity=1 ] (248,105.64) .. controls (248,92.73) and (258.59,82.27) .. (271.65,82.27) .. controls (284.71,82.27) and (295.3,92.73) .. (295.3,105.64) .. controls (295.3,118.54) and (284.71,129) .. (271.65,129) .. controls (258.59,129) and (248,118.54) .. (248,105.64) -- cycle ;
\draw  [color={rgb, 255:red, 0; green, 0; blue, 0 }  ,draw opacity=0 ][fill={rgb, 255:red, 23; green, 76; blue, 208 }  ,fill opacity=1 ] (319,170.64) .. controls (319,157.73) and (329.59,147.27) .. (342.65,147.27) .. controls (355.71,147.27) and (366.3,157.73) .. (366.3,170.64) .. controls (366.3,183.54) and (355.71,194) .. (342.65,194) .. controls (329.59,194) and (319,183.54) .. (319,170.64) -- cycle ;
\draw    (357.9,59.84) -- (394.9,91.84) ;

\draw    (286.9,123.84) -- (323.9,155.84) ;

\draw    (361,155.28) -- (403.9,122.84) ;

\draw    (289,90.28) -- (325.05,60.3) ;

\draw    (426.9,121.84) -- (439.9,184.84) ;

\draw  [color={rgb, 255:red, 0; green, 0; blue, 0 }  ,draw opacity=0 ][fill={rgb, 255:red, 196; green, 11; blue, 18 }  ,fill opacity=1 ] (420.9,204.84) .. controls (420.9,191.94) and (431.49,181.48) .. (444.55,181.48) .. controls (457.61,181.48) and (468.2,191.94) .. (468.2,204.84) .. controls (468.2,217.74) and (457.61,228.2) .. (444.55,228.2) .. controls (431.49,228.2) and (420.9,217.74) .. (420.9,204.84) -- cycle ;
\draw  [color={rgb, 255:red, 0; green, 0; blue, 0 }  ,draw opacity=0 ][fill={rgb, 255:red, 196; green, 11; blue, 18 }  ,fill opacity=1 ] (220.3,233.64) .. controls (220.3,220.73) and (230.89,210.27) .. (243.95,210.27) .. controls (257.01,210.27) and (267.6,220.73) .. (267.6,233.64) .. controls (267.6,246.54) and (257.01,257) .. (243.95,257) .. controls (230.89,257) and (220.3,246.54) .. (220.3,233.64) -- cycle ;
\draw  [color={rgb, 255:red, 0; green, 0; blue, 0 }  ,draw opacity=0 ][fill={rgb, 255:red, 196; green, 11; blue, 18 }  ,fill opacity=1 ] (421.3,29.64) .. controls (421.3,16.73) and (431.89,6.27) .. (444.95,6.27) .. controls (458.01,6.27) and (468.6,16.73) .. (468.6,29.64) .. controls (468.6,42.54) and (458.01,53) .. (444.95,53) .. controls (431.89,53) and (421.3,42.54) .. (421.3,29.64) -- cycle ;
\draw  [color={rgb, 255:red, 0; green, 0; blue, 0 }  ,draw opacity=0 ][fill={rgb, 255:red, 196; green, 11; blue, 18 }  ,fill opacity=1 ] (157.3,61.64) .. controls (157.3,48.73) and (167.89,38.27) .. (180.95,38.27) .. controls (194.01,38.27) and (204.6,48.73) .. (204.6,61.64) .. controls (204.6,74.54) and (194.01,85) .. (180.95,85) .. controls (167.89,85) and (157.3,74.54) .. (157.3,61.64) -- cycle ;
\draw    (202.9,71.84) -- (250.9,96.84) ;

\draw    (362.9,35.84) -- (421.3,29.64) ;

\draw    (322.9,183.84) -- (265.9,225.84) ;

\draw    (341.65,68) -- (342.65,147.27) ;

\draw    (294.1,107.54) -- (391.9,108.84) ;

\end{tikzpicture}

\caption{An example of a graph containing a clique. The blue nodes represent a fully connected subgraph of dimension 4, whereas the red nodes do not belong to the clique.}
\label{fig:clique}
\end{figure}

\paragraph{\textbf{Clique localization}}
A clique is a complete graph, i.e. a graph in which each node is connected with all the others. In a network, overlapping cliques (i.e. cliques that share some nodes) are admitted. 
Clique localization is a particular instance of the subgraph matching problem, with $S$ being complete. However, the several symmetries contained in a clique makes the graph isomorphism test more difficult. Indeed, it is known that the graph isomorphism has polynomial time solutions only in absence of symmetries.
A clique example is shown in Fig. \ref{fig:clique}.
In the experiments, we consider a dataset composed by graphs having 7 nodes each, where the dimension of the maximal clique is 3 nodes.

We designed a batch of experiments on these two tasks aimed at validating our simple local optimization approach to constraint-based networks. In particular, we want to show that our optimization scheme can learn better transition and output functions than the corresponding GNN of \cite{DBLP:journals/tnn/ScarselliGTHM09}. Moreover, we want to investigate the behaviour of the algorithm for different choices of the function $\mathcal{G}(x)$, i.e. when changing how we enforce the state convergence constraints. In particular, we tested functions with different properties: \textit{$\epsilon$-insensitive functions}, i.e $\mathcal{G}(x)=0,\  \forall x: -\epsilon \le x \le \epsilon$,  \textit{unilateral functions}, i.e. $\mathcal{G}(x) \in \mathbb{R}^+$, and \textit{bilateral functions}, i.e. $\mathcal{G}(x) \in \mathbb{R}$ (a $\mathcal{G}$ function is either unilateral or bilateral). The considered functions are shown in Table~\ref{tab:g_functions}. 

Following the experimental setting of \cite{DBLP:journals/tnn/ScarselliGTHM09}, we exploited a training, validation and test set having the same size, i.e. 100 graphs each. We tuned the hyperparameters on the validation data, by selecting the node state dimension from the set  $\{5, 10, 35\}$,  the dropout drop-rate from the set $ \{0., 0.7 \}$,  the state \textit{transition function} from $\{f_{a,v}^{\text{(AVG)}}, f_{a,v}^{\text{(SUM)}}\}$ and their number of hidden units from  $ \{5, 20, 50\}$. We used the Adam optimizer (TensorFlow). Learning rate for parameters $\theta_{f_a}$ and $\theta_{f_r}$ is selected from the set $\{10^{-5}, 10^{-4},10^{-3}\}$, and the learning rate for the variables $x_v$ and $\lambda_v$ from the set $ \{ 10^{-4}, 10^{-3}, 10^{-2}\}$.

We compared our model with the equivalent GNN in \cite{DBLP:journals/tnn/ScarselliGTHM09}, with the same number of hidden neurons of the $f_a$ and $f_r$ functions. For the comparison, we exploited the GNN Tensorflow implementation \footnote{The framework is available at \url{https://github.com/mtiezzi/gnn}. The documentation is available at \url{http://sailab.diism.unisi.it/gnn/}} introduced in \cite{rossi2018inductive}.
Results are presented in Table~\ref{tab:artificial_results}.

\begin{table*}[h!]\
\centering

\begin{tabular}{@{}lccccccc@{}}\toprule
 Datasets &  {\textsc{IMDB-B}} & {\textsc{IMDB-M}}  & {\textsc{MUTAG}} &  {\textsc{PROT.}}  & {\textsc{PTC}} & {\textsc{NCI1}}  \\
 \text{\# graphs }  & 1000  & 1500  &  188 & 1113 & 344  &  4110     \\
 \text{\# classes }   &  2  & 3  &   2 & 2  &  2   & 2 \\
 \text{Avg \# nodes }  &  19.8   & 13.0  &  17.9 & 39.1  & 25.5  &  29.8  
\\ \midrule
 \textsc{DCNN}             &  49.1 & 33.5     &   67.0 & 61.3 & 56.6 &  62.6 \\
 \textsc{PatchySan}        & 71.0 $\pm$ 2.2  & 45.2 $\pm$ 2.8  & { 92.6 $\pm$ 4.2}  & 75.9 $\pm$ 2.8  & 60.0 $\pm$ 4.8   &  78.6 $\pm$ 1.9     \\ 
 \textsc{DGCNN}           & 70.0 &  47.8     & 85.8        &  75.5   & 58.6  & 74.4   \\ 
 \textsc{AWL}               & 74.5 $\pm$ 5.9  & 51.5 $\pm$ 3.6    & 87.9 $\pm$ 9.8        & -- & --  & --  \\
 \textsc{GIN}   &  { 75.1 $\pm$ 5.1}     & { 52.3 $\pm$ 2.8}    &  { 89.4 $\pm$ 5.6}     & { 76.2 $\pm$ 2.8}    & { 64.6 $\pm$ 7.0}  & { 82.7 $\pm$ 1.7}\\
  \textsc{GNN} & 60.9  $\pm$ 5.7  & 41.1  $\pm$ 3.8 & 88.8  $\pm$ 11.5 & 76.4  $\pm$ 4.4 & 61.2  $\pm$ 8.5 & 51.5  $\pm$ 2.6\\
  \textsc{LP-GNN*} & 71.2  $\pm$ 4.7 &  46.6  $\pm$ 3.7 & 90.5  $\pm$ 7.0 & 77.1  $\pm$ 4.3  & 64.4  $\pm$ 5.9 & 68.4  $\pm$ 2.1\\
\bottomrule
\end{tabular}

 \caption{We report the average accuracies and standard deviations for the graph classification benchmarks, evaluated on the test set, and we compare multiple GNN models. The proposed model is denoted as LP-GNN and marked with a star. Even though it exploits only shallow representation of nodes, our model performs, on average, on-par to other top models, setting a new state-of-the-art for the Proteins dataset. 
 }
 \label{tab:results}
\end{table*}

Constraints characterized by \textit{unilateral functions} usually offer better performances than equivalent bilateral constraints. This might be due to the fact that keeping constraints positive (as in unilateral constraints) provides a more stable learning process. Moreover, smoother constraints (i.e \textit{squared}) or $\eps$-insensitive constraints tend to perform slightly better than the hard versions. This can be due to the fact that as the constraints move closer to 0 they tend to give a small or null contribution, for \textit{squared} and \textit{abs-$\eps$} respectively, acting as regularizers. 


\subsection{Graph Classification}

We used 6 graph classification benchmarks: 4 bioinformatics datasets (MUTAG, PTC, NCI1, PROTEINS) and 2 social network datasets (IMDB-BINARY, IMDB-MULTI)~\citep{yanardag2015deep}, which are becoming popular for benchmarking GNN models. In the bioinformatic graphs, the nodes have categorical input labels (e.g. atom symbol). In the social networks, there are no input node labels. In this case, we followed what has been recently proposed in \cite{DBLP:journals/corr/abs-1810-00826}, i.e. using one-hot encodings of node degrees. Dataset statistics are summarized in Table \ref{tab:results}.

We compared the proposed Lagrangian Propagation GNN (LP-GNN) scheme with some of the state-of-the-art neural models for graph classification, such as Graph Convolutional Neural Networks. All the GNN-like models have a number of layers/iterations equal to 5. An important difference with these models is that, by using a different transition function at each iteration, at a cost of a much larger number of parameters, they have a much higher representational power. Even though our model could, in principle, stack multiple diffusion processes at different levels (i.e. different latent representation of the nodes) and, then, have multiple transition functions, we have not explored this direction in this paper.  In particular, the models used in the comparison are: 
Diffusion-Convolutional Neural Networks (DCNN)~\citep{DBLP:conf/nips/AtwoodT16}, PATCHY-SAN \citep{DBLP:conf/icml/NiepertAK16}, Deep Graph CNN (DGCNN) \citep{DBLP:conf/aaai/ZhangCNC18}, AWL \citep{ivanov2018anonymous} , GIN-GNN \citep{DBLP:journals/corr/abs-1810-00826}, original GNN \citep{DBLP:journals/tnn/ScarselliGTHM09}.  Apart from original GNN, we report the accuracy as reported in the referred papers.

We followed the evaluation settings in \citep{DBLP:conf/icml/NiepertAK16}. In particular, we performed 10-fold cross-validation and reported both the average and standard deviation of validation accuracies across the 10 folds within the cross-validation. The stopping epoch is selected as the epoch with the best cross-validation accuracy averaged over the 10 folds. 
We tuned the hyperparameters by searching: (1) the number of hidden units for both the $f_a$ and $f_r$ functions from the set $\{5,20,50, 70, 150\}$; (2)  the state transition function from  $\{f_{a,v}^{\text{(AVG)}}, f_{a,v}^{\text{(SUM)}}\}$; (3) the dropout ratio from $\{0, 0.7\}$; (4) the size of the node state $x_v$ from $\{ 10, 35, 50, 70, 150\}$; (5) learning rates for both the $\theta_{f_a}$, $\theta_{f_r}$, $x_v$ and $\lambda_v$ from $\{ 0.1, 0.01, 0.001 \}$. Results are shown in Table \ref{tab:results}.

As previously stated, differently from the baseline models, our approach does not rely on a deep stack of layers based on differently learnable filters.
Despite of this fact, the simple GNN model trained by the proposed scheme offers performances that, on average, are preferable or on-par to the ones obtained by more complex models that exploit a larger amount of parameters.

Moreover, it is interesting to note that for current GNN models, the role of the architecture depth is twofold. First, as it is common in deep learning, depth is used to perform a multi-layer feature extraction of node inputs. Secondly, it allows node information to flow through the graph fostering the realisation of a diffusion mechanism. Conversely, our model strictly splits these two processes. We believe this distinction to be a fundamental ingredient for a clearer understanding of which mechanism, between diffusion and node deep representation, is concurring in achieving specific performances. Indeed, in this paper, we show that the diffusion mechanism paired only with a simple shallow representation of nodes is sufficient to match performances of much deeper and complex networks.

\section{Conclusions and Future Work}
\label{sec:concl}
We showed that formulation of the GNN learning task as a constrained optimization problem allows us to avoid the explicit computation of the fixed point needed to encode the graph. The proposed framework defines how to jointly optimize the model weights and the state representation without the need of separate phases. This approach simplifies the computational scheme of GNNs and allows us to incorporate alternative strategies in the fixed point optimization by the choice of the constraint function ${\cal G}()$. As shown in the experimental evaluation, the appropriate functions may affect generalization and robustness to noise.

Future work will be devoted to explore systematically the properties of the proposed algorithm in terms of convergence and complexity. Moreover, we plan to extend the experimental evaluation to verify the algorithm behaviour with respect to either the characteristics of the input graphs, such as the graph diameter, the variability in the node degrees, the type of node and arc features or to the model architecture (f.i. type of the state transition function, of the constraint function, etc.). Furthermore, the proposed constraint-based scheme can be extended to all the other methods proposed in the literature that exploit more sophisticated architectures. 

Finally, LP-GNN can be extended allowing the diffusion mechanism to take place at multiple layers allowing a \textit{controlled} integration of diffusion and deep feature extraction mechanisms.

\section*{ Acknowledgments}
This work was partly supported by the PRIN 2017 project RexLearn, funded by the Italian Ministry of Education, University and Research (grant no. 2017TWNMH2). 

\bibliography{refe}

\end{document}